\documentclass[a4paper, 10pt, conference]{ieeeconf}

\IEEEoverridecommandlockouts                              
\usepackage[utf8]{inputenc}
\usepackage[T1]{fontenc}
\usepackage{cite}
\usepackage{amsmath,amssymb,amsfonts}
\usepackage{algorithmic}
\usepackage{graphicx}
\usepackage{textcomp}
\usepackage{xcolor}
\usepackage{breqn}
\usepackage{textcase}
\usepackage[export]{adjustbox}
\usepackage{subcaption}

\title{\LARGE \bf Efficient statistical validation with edge cases to evaluate Highly Automated Vehicles}

\author{Dhanoop Karunakaran$^{1}$, Stewart Worrall$^{1}$, Eduardo Nebot$^{1}$
\thanks{$^{1}$D.Karunakaran, S. Worrall,  E. Nebot  are with the Australian Centre for Field Robotics (ACFR) at the University of Sydney (NSW, Australia).
       E-mails: {\tt\small \{d.karunakaran,  s.worrall,  e.nebot\}@acfr.usyd.edu.au}}
}

\begin{document}

\maketitle
\thispagestyle{empty}
\pagestyle{empty}

\begin{abstract}
The widescale deployment of Autonomous Vehicles (AV) seems to be imminent despite many safety challenges that are yet to be resolved. 
It is well known that there are no universally agreed Verification and Validation (VV) methodologies to guarantee absolute safety, which is crucial for the acceptance of this technology. 
Existing standards focus on deterministic processes where the validation requires only a set of test cases that cover the requirements.
Modern autonomous vehicles will undoubtedly include machine learning and probabilistic techniques that require a much more comprehensive testing regime due to the non-deterministic nature of the operating design domain.
A rigourous statistical validation process is an essential component required to address this challenge. 
Most research in this area focuses on evaluating system performance in large scale real-world data gathering exercises (number of miles travelled), or randomised test scenarios in simulation. 

This paper presents a new approach to compute the statistical characteristics of a system's behaviour by biasing automatically generated test cases towards the worst case scenarios, identifying potential unsafe edge cases.
We use reinforcement learning (RL) to learn the behaviours of simulated actors that cause unsafe behaviour measured by the well established RSS safety metric.
We demonstrate that by using the method we can more efficiently validate a system using a smaller number of test cases by focusing the simulation towards the worst case scenario, generating edge cases that correspond to unsafe situations.
\end{abstract}

\section{Introduction}
\begin{figure}[t]
\includegraphics[width=0.98\columnwidth]{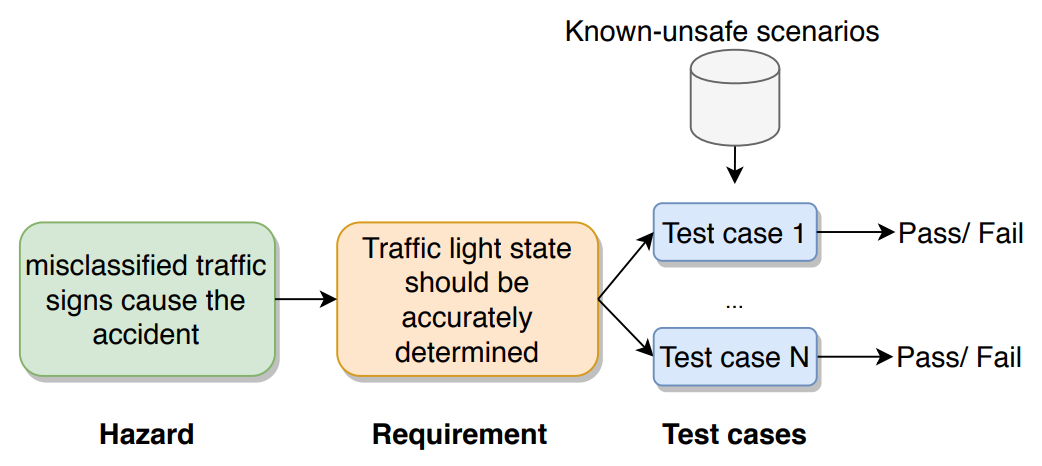}
\caption{An example of traditional validation approach for traffic light classifier based on ISO26262}
\label{fig:requirements}
\end{figure}

The non-deterministic nature of the urban road environment is a significant challenge for the validation of Highly Automated Vehicles(HAV) \cite{koopman2016challenges}\cite{testing_autonomous_philipp_helle}\cite{non_deter_world}. 
It is difficult for prediction and classification algorithms to anticipate the behavior of road participants, and to handle changes in conditions such as lighting, weather and other environmental conditions. It is further complicated by the use of machine learning and probabilistic methods. Figure~\ref{fig:requirements} depicts an example of a traditional validation approach for traffic light classifier based on ISO26262. The traditional deterministic approach evaluates the system by verifying a number of test cases, derived from known-unsafe scenarios, as per the requirements\cite{iso26262}\cite{khan2017iso}. 
When non-determinism is present, it creates unknown-unsafe scenarios or edge cases that may not be accounted for during the validation stage. 
This leads to the necessity for an efficient approach to validate such systems. 
When non-determinism is present, the evaluation process should be focused more on estimating the statistical credibility of the system performance\cite{koopman2016challenges} by testing it over a significant number of test cases. For example, determining whether the false detection rate is higher than a defined rate as a measure\cite{koopman2016challenges}. We define an edge case as an unknown-unsafe scenario, which is difficult to predict using existing deterministic testing methodologies, potentially leading to accidents when the system is deployed on public roads. For example, the failure to detect an object that results in a critical situation when operating on public roads is an edge case. The scenario is defined as a sequence of events, and each event is a collection of actions and maneuvers executed by the ego vehicle and other traffic participants\cite{elrofai2016scenario}.

\begin{figure}[t]
\includegraphics[width=0.98\columnwidth]{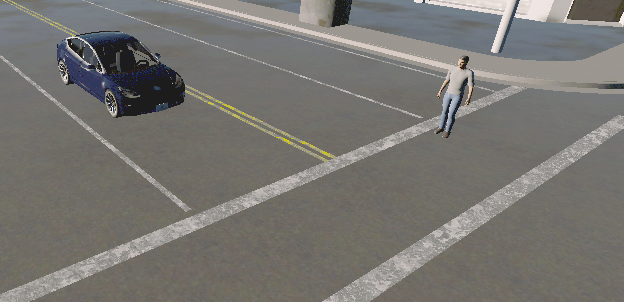}
\caption{Carla simulation of the experiment: a pedestrian crossing the street while the vehicle is moving. }
\label{fig:ped_cross_sample}
\end{figure}

One of the validation methodologies for HAV involves computing the statistical characteristics representing the behavior of the system. To a certain extent, this method is an effective approach to handle non-determinism in testing. Still, much of the research in this space has been focused on computing the statistical characteristics of the behavior based on randomly generated test cases, or through on-road testing. There is, however, no assurance with either of these approaches that the tests will cover the most challenging, or edge cases. As edge cases are rare and randomly distributed, most of the distance traveled will not contain edge cases or unknown-unsafe scenarios using on-road data for the statistical validation\cite{amersbach2019functional_parameter_space_explosion}\cite{klamann2019defining}.

The contribution of this paper is the proposal of a method to compute the statistical characteristics of a system's behavior by biasing automatically generated test cases towards the edge cases. This method reduces the need for validation over a large number of generated scenarios or billion of miles of real-world data. Reinforcement learning (RL) has been used in other fields of research to optimize the learning of an action set towards failure scenarios or edge cases that maximize the reward\cite{adaptive_stress_testing_flight}. In our work, RL generates scenarios during the learning phase that trend towards the edge cases, enabling the final model to test the system with more challenging cases. Each generated scenario is validated against the specifications to compute the probability that the system can satisfy the specifications. In this experiment, we are using a well established Responsibility-Sensitive Safety (RSS) as a safety metric to evaluate the system. The pedestrian crossing illustrated in figure~\ref{fig:ped_cross_sample} is used as the experimental context in the simulator to test our method. 

The remainder of the paper is organised as follows. Section  II presents related work, followed by methodologies. Section IV describes the proposed experiments, followed by discussion and concluding remarks.

\section{Related work}

The current motivation of public road testing is to collect miles of data to identify the edge cases\cite{koopman2018toward}. As edge cases are inherently rare, this approach requires the collection of a significant number of miles (billions) in order to observe rare events, which is time consuming and expensive\cite{mobileye_rss}\cite{amersbach2019defining}. On top of that, the accumulated evidence will be undermined with every software change\cite{mobileye_rss}. As these challenging cases are rare and randomly distributed, most of the distance covered will have no effect on the validation\cite{amersbach2019functional_parameter_space_explosion}\cite{klamann2019defining}. 

To the best of our knowledge, research in the statistical validation of HAV systems focuses on computing the probability of a system satisfying the specifications by generating random test cases without much emphasis on the edge cases. In \cite{validation_non_deterministic}, Statistical Model Checking (SMC) is used to compute the statistical characteristics for the behaviour of the system. They focus on calculating the probability of system failure by analysing the result of randomly generated test cases in simulation without much focus on unknown-unsafe cases or edge cases. Similarly, in \cite{de2017assessment_smc}, the authors proposed generating the new test cases in simulation to estimate the performance of ADAS from recorded real-life scenarios. Nevertheless, they still rely on data collected from public roads which is time-consuming and expensive. Also, there is no guarantee that the collected data will include edge cases.

In our work, we present an efficient statistical validation method for analyzing the behavior of the system. This is achieved by evaluating a safety metric over different generated scenarios using reinforcement learning to find the parameters corresponding to the worst case scenarios. The proposed method reduces the need for billions of miles of data and large randomly generated test sets by biasing automatically generated scenarios towards the edge cases.

\section{Methodologies}
This section provides the background knowledge and details about proposed method.

\subsection{Markov Decision Processes and Reinforcement Learning}

 A Markov Decision Process (MDP) is a Markov Reward Process (MRP) with decision capabilities. It represents an environment in which all of the states hold the Markov  property\footnote{Markov property is defined as "future is independent of past, given present"\cite{rl_intro_sutton}.}\cite{rl_palanisamy}. The solution of a MDP is an optimal policy that evaluates the best action to choose from each state\cite{rl_another_book}. In MDP, at each time step$\ t$, the agent takes action $\ a_t$ in the environment and receives the observation from the environment in the form of a new state$\ s_{t+1}$ given the current state$\ s_t$. A state is the information used to determine what happens next\cite{rl_palanisamy}. The agent and environment have their own state representation. State$\ s_t$ is the state representation of the agent constructed by either directly or indirectly observing the environment state\cite{rl_palanisamy}. Reward$\ r_t$ is a scalar quantity to give feedback to the agent about how good was the action$\ a_t$ based on the state$\ s_t$ from the environment. 
 
 A Reinforcement Learning(RL) algorithm finds an optimal policy that maximizes the reward by interacting with the environment that is modeled as an MDP where no prior knowledge about the MDP is available\cite{rl_another_book2}. A policy is defined as a mapping from an agent state to an action in order to move to the next state. A value function is an important aspect of RL to determine the optimal policy.  There are two types of value functions: state value and action value.

An off-policy TD control RL algorithm, known as Q Learning, is used in the proposed framework. This algorithm is concerned with finding the optimal action value$\ Q^*$ which gives the maximum expected return, starting from a state $\ s_t$, taking action $\ a_t$, and then following the optimal policy from the next state $\ s_{t+1}$ onward\cite{david_silver_rl}. The Bellman equation is for the optimal action value$\ Q^* $ as follows\cite{david_silver_rl}\cite{deeprlped}:

\begin{equation} \label{eq1}
Q^*(s, a) = E\left[R+\gamma max_{a_{t+1}}Q^*(s_{t+1}, a_{t+1})|s,a \right]
\end{equation}

It enables the agent to learn the long-term value of taking action$\ a_t$ in state$\ s_t$ so that the agent can take actions that will maximise the expected reward\cite{rl_palanisamy}\cite{david_silver_rl}. The equation for the one-step Q-learning is as follows\cite{rl_intro_sutton}:

\begin{multline}
Q(s_t,a_t) \leftarrow Q(s_t,a_t)+\\
\alpha\left[r_{t+1}+\gamma max(Q(s_{t+1},a_t))-Q(s_t,a_t)\right]
\end{multline}

Based on this equation, the learned action value function $\ Q$ approximates the optimal action-value function$\ Q^*$ regardless of the policy being followed\cite{rl_intro_sutton}\cite{david_silver_rl}. 

\subsection{Deep Q Network (DQN)}
Deep Q Network replaces the use of a Q-table to store the Q values, with the neural network weights. This is important because the original Q-table is not scalable when there is a large set of state-action pairs \cite{deeprlped}. Deep Learning (DL) is based on minimising the error estimated by the loss function by optimising the weights$\ \theta$. Error or loss is measured as the difference between the predicted and actual result. In a DQN, we can represent our loss function as a squared error of the target Q value and prediction Q value
\cite{david_silver_rl}\cite{deeprlped}:

\begin{multline}
L(\theta)=[((r+\gamma max_{a_{t+1}} \\
Q(s_{t+1}, a_{t+1};\theta^{target}))-Q(s,a;\theta^{pred}))^2] 
\end{multline}

\begin{figure}[h]
\includegraphics[width=0.95\columnwidth]{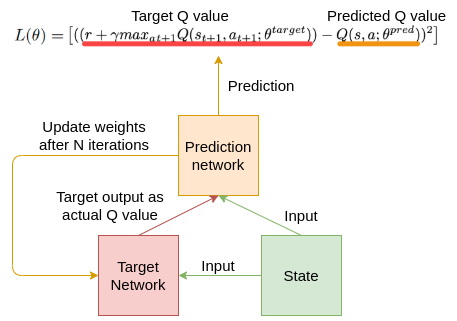}
\caption{DQN Architecture}
\label{fig:dqn_architecture}
\end{figure}

As shown in figure~\ref{fig:dqn_architecture}\cite{deeprl_analytics}, there are two separate networks in a DQN with same architecture to estimate the target and prediction Q values for the stability of the Q-learning algorithm. Also, the past experience is stored in memory known as replay memory. At each time step$\ t$, a subset of replay memory is randomly selected and fed through the networks to minimise $\ L(\theta)$ and to optimise the weights\cite{david_silver_rl}\cite{deeprlped}.

\subsection{Responsibility Sensitive Safety (RSS)}
The ultimate goal of the formal method is to guarantee that an agent will not cause an accident rather than to guarantee an agent will not be involved in an accident\cite{mobileye_rss}. RSS has formalized the set of rules using four realms: Safe Distance, Dangerous Situation, Proper Response, and Responsibility\cite{rss}. A safe distance is calculated longitudinally and laterally based on the formula provided by the method. This method considers the worst-case scenarios which eliminates the need for estimating road user intentions. In this paper, we focus on the longitudinal safe distance. By definition, the longitudinal safe distance is the minimum distance required for the ego vehicle to stop in time if the vehicles or objects in front brake abruptly.

\begin{equation} \label{eq1}
d_{min}=\left[v_r\rho+\frac{1}{2}a_{max,a}\rho^2+\frac{(v_r+\rho a_{max,a})^2}{2a_{min,b}}-\frac{v_f^2}{2a_{max,b}}\right]
\end{equation}

where $\ d_{min} $ represents the longitudinal safe distance, $\ v_r $  and $\ v_f $ are the velocity of the agent vehicle and front vehicle, respectively. $\ a_{min,b}$ is the minimum reasonable braking force of the agent vehicle, $\ a_{max ,b}$ is the maximum braking force of the front vehicle. In terms of acceleration, $\ a_{max,a} $ is the maximum acceleration of the front vehicle. $\rho$ is the agent vehicle response time.

Based on the safe distance calculation, RSS can determine whether or not the ego vehicle is in a dangerous situation. As per the RSS rule, the proper response should be determined and executed once it is determined that the vehicle is in a dangerous situation. If there is a collision, the responsibility of the accident can be assigned to the party that did not comply with the proper response.

\subsection{Proposed Method}

When non-determinism is present, estimating the statistical characteristics of a system's behavior will be an effective approach \cite{validation_non_deterministic}. As discussed, most approaches aim to compute the statistical characteristics without much focus on edge cases. The proposed method addresses the aspect of statistical validation by biasing a scenario towards the edge cases.

\begin{figure}[h]
\includegraphics[width=0.95\columnwidth]{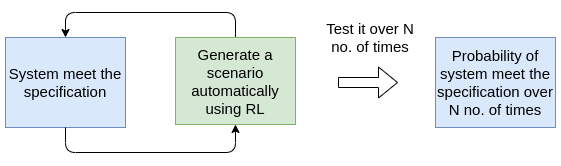}
\caption{Proposed method}
\label{fig:figure_general_method}
\end{figure}

The overview of the proposed method is as depicted in Figure~\ref{fig:figure_general_method}. RL is a type of machine learning algorithm that enables us to generate scenarios. Each scenario is then validated against the specification to build the statistical figure. The specification can vary based on the experiment. We set the RL agent to learn towards the failure scenarios by rewarding the agent for taking actions that lead to failure as measured by the RSS metric. The optimal solution of the agent will generate failure scenarios that have the maximum reward. We define a scenario as a failure scenario when the system does not satisfy the specifications.
Similarly, a success scenario defined as a system that satisfies the specification. Success and failure scenarios are generated using RL during the learning phase. We consider an episode of a RL agent as a scenario. In each episode, RL explores a different solution to maximize the reward. The advantage of using this method is that we can validate the system with randomly generated test cases during the early training phase, then towards the end of the learning phase we validate with more challenging scenarios. Statistical characteristics of the system behavior is computed by finding the probability of the system satisfying the specifications. This is done by dividing the total number of success scenarios with the total sum of scenarios, as shown below:
\begin{equation} 
p_r = \frac{Success_{scenarios}}{Total_{scenarios}}
\end{equation}
Failure scenarios that have maximum reward are most likely the edge cases that were not accounted for during the development of the system. We argue that the method of automatically generating scenarios and optimizing the learning towards finding out the edge cases is a better approach than testing over a large number of randomly generated test cases. The proposed approach is an effective method that enables us to incorporate the edge cases in statistical validation of HAV within an Operational Design Domain(ODD).

\section{Experiment}
\begin{figure}[h]
\includegraphics[width=0.95\columnwidth]{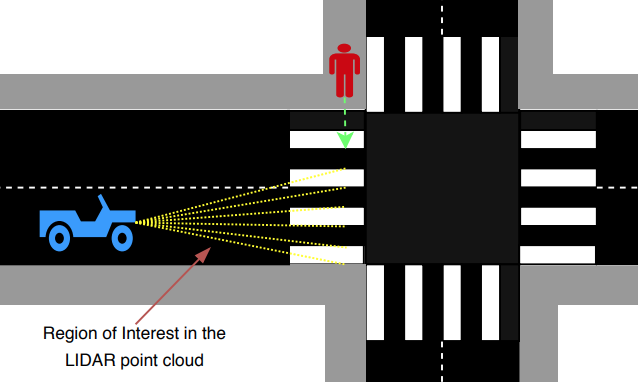}
\caption{Pedestrian crossing intersection}
\label{fig:ped_crossing_intersection}
\end{figure}
\begin{figure}[h]
\includegraphics[width=0.95\columnwidth]{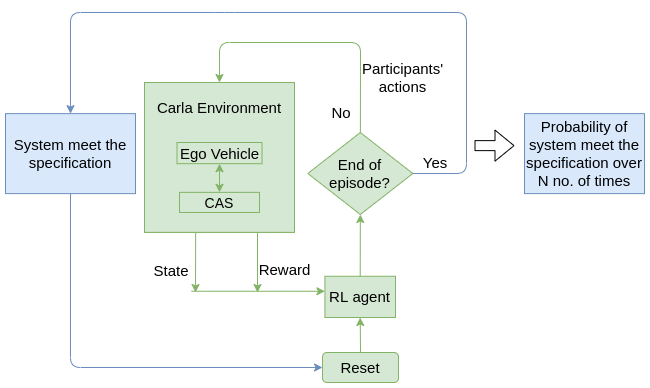}
\caption{Implementation of proposed method}
\label{fig:figure_proposed_method}
\end{figure}

 
In this work we conduct experiments at a pedestrian crossing using the Carla simulator as shown in Figure~\ref{fig:ped_crossing_intersection}. We have implemented the proposed validation method, as depicted in Figure~\ref{fig:figure_proposed_method}. The goal of this experiment is to create scenarios where we control the sequence of simluated pedestrian's actions to validate the system under test (SUT).
These actions are represented as a sequence of time-stamped pedestrian speed values.  Each sequence of actions is considered as one scenario/episode. They are generated such that the RL agent performs training towards learning the failure scenarios as the optimal solution. In each scenario, a pedestrian is walking through the crosswalk while the ego vehicle is approaching the crosswalk. The proposed method enables us to validate the system against the defined specification along with edge cases to generate the probability measure. The method enables us to validate the system with a smaller number of scenarios compared to using only randomised inputs. As we exercise the RL to bias the inputs towards the failure scenarios, the statistical validation with edge cases can be computed. We argue that this is a better method for validation when non-determinism is present.  

We have selected a relatively simple task in order to demonstrate the method, with more complexity to be added in future work. We expect that this approach can be used to determine the parameters of many autonomous vehicle sub-systems such as collision avoidance systems, and to be able to demonstrate safety when working under a particular ODD.  At this stage, we are not considering the pedestrian intention and uncertainties from the ego vehicle's sensor measurement. The randomness of the process is introduced with the initial position and speed of the pedestrian. There are two initial positions where a pedestrian can start and 40 different possible speeds for the pedestrian to walk.  We also introduce randomness by adding noise to vary the speed of ego vehicle.

The main components of the experiment are explained below, before discussing the implementation of the proposed method.

\begin{figure}[h]
\includegraphics[width=0.95\columnwidth]{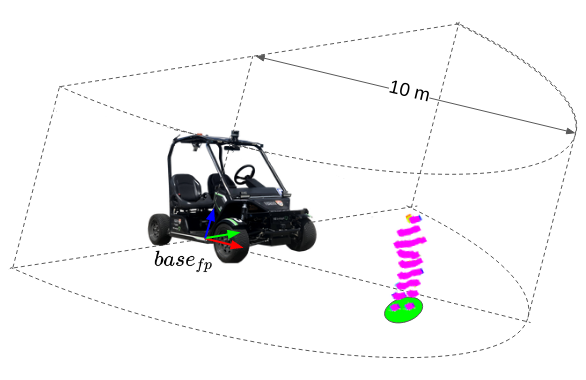}
\caption{Collision Avoidance System (CAS)'s detection range }
\label{fig:sut}
\end{figure}

\textbf{System Under Test (SUT)}: We are using a simple Collision Avoidance System (CAS) provided by the Carla simulator to test our method. As depicted in the Figure~\ref{fig:sut}, the maximum detection range of the system is 10 meters. The systems basic functionality is to control the speed of the vehicle to avoid a collision when an object is detected. Importantly, we are treating this system as a blackbox system.  

\textbf{Environment}: We use the Carla simulator\cite{carla_simulator} to create the environment for RL where the SUT is connected to the ego vehicle. An episode is terminated when the ego vehicle travels further than 40m, the episode runs for more than 100 seconds, or the ego vehicle is involved in a collision.

\textbf{Action space}: The action space contains 40 actions comprised of different possible speeds for the pedestrian at each time step, ranging from 0 to 10 m/s as shown in Table~\ref{table:action_space}. 

\begin{table}[h!]
\centering
 \caption{Action space}
 \label{table:action_space}
 \begin{tabular}{||c c||} 
 \hline
 Action & Speed \\ [0.5ex] 
 \hline\hline
 0 & 0.00 m/s\\ 
 1 & 0.25 m/s\\
 2 & 0.50 m/s\\
 ...&...\\
 ...&...\\
 38 & 9.50 m/s\\
 39 & 9.75 m/s\\
 40 & 10.00 m/s\\ [1ex] 
 \hline
 \end{tabular}
\end{table}

\textbf{State space}: State space consists of the relative speed and euclidean distance of the pedestrian w.r.t the ego vehicle. The speed is expressed in meters per second and the distance is in meters. 
\begin{equation}
S = (s_{ped1-ego},dist_{ped1-ego})
\end{equation}

\textbf{Reward function}: This function determines how good the action is for a given state. 
The reward function is calculated as the value measured by the RSS safety metric.
At each time steps$\ t $, the minimum safe distance is measured using the RSS formula. This metric is compared with the Euclidean distance to the pedestrian. The normal behavior of the SUT is to reduce the speed of the ego vehicle when an object is detected. If the Euclidean distance is less than the safe distance measured by RSS, it is considered an improper response from the SUT. This situation will produce timesteps marked as a failure, though as the primary goal of the RL agent is to learn the challenging cases, the reward is set to a value of 2, biasing the RL to this condition. If the Euclidean distance is greater than the safe distance, then it is considered a proper response and the reward will be -2.  As shown in the Figure~\ref{fig:sut}, the vehicle detects the pedestrian only within the region of interest. Zero reward is given for the rest of the time. Also, zero reward is allocated if the ego vehicle is involved in an accident. The reward function is summarised as below:
\begin{equation}
R = \Bigg\{\begin{matrix}2& Distance_{Euclidean} < Distance_{RSS}\\-2&Distance_{Euclidean} >= Distance_{RSS}\\
0& collision/non\ region\ of\ interest
\end{matrix}
\end{equation}

\textbf{Episode}: It is a sequence of states and actions. Each episode is considered as a scenario for statistical validation. One episode typically contains 40 to 55 timesteps. The end of an episode is determined by time, distance traveled, and in the event of a collision. The agent is set to run up to 10000 episodes. We set this number arbitrarily to demonstrate the proposed method.

\textbf{DQN architecture and training}: The DQN is implemented using two neural networks as depicted in Figure~\ref{fig:dqn_architecture}. The result from the target model is treated as the ground truth for the prediction network. At each time step, a queue called replay memory stores a set of values generated by the RL agent action:$\ state, action, reward, newstate, done$. In this case, $\ done$ determines whether or not an the episode has concluded. Currently, the maximum storage limit of the queue is set to 2000. We select a batch size of 32 items. One epoch of training for the prediction network is completed at each timestep, and this process repeats for every timestep. The queue enables the training of the network for the batch size and random dataset selection. The target network weights are updated with weights from the prediction network at every 25 episodes. 

\begin{figure}[h]
\includegraphics[width=0.95\columnwidth, trim={3cm 0 3cm 0},clip ]{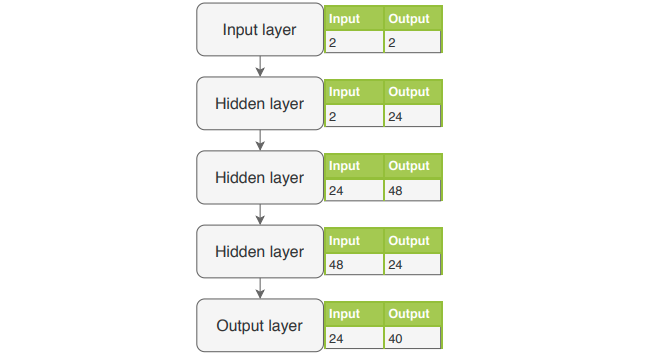}
\caption{DL model architecture}
\label{fig:model}
\end{figure}
As shown in Figure~\ref{fig:model}, both DL models are four layer networks. The first layer contains two neurons to accept the two states and 40 neurons in the last layer to represent the action space. We are using the Adam optimisation algorithm with learning rate of 0.01.

In RL, the balance between exploitation vs exploration is an important aspect. 
Exploration at the early stages enables the model not to get stuck in a local maxima predicted by the relatively untrained model.
Initially, an agent is weighted towards exploration compared to exploitation as determined by the $\ \epsilon$ value. The initial value of $\ \epsilon$ is set to 1, decaying at each time step by multiplying by 0.995 until $\ \epsilon_{min}$=0.001. During this time, the actions selected for the agent will be random. 

A DQN generally requires a large number of episodes to converge on an optimal set of action values $\ Q^* $. This will depend on the size of the state space and the action space. 

\textbf{Evaluation metric}: Is a metric to identify whether SUT meets/satisfies the specifications. In this experiment, our specification depends on$\ distance_{Euclidean}$ and$\ distance_{RSS}$.  The$\ distance_{Euclidean}$ is the Euclidean distance between pedestrian and ego vehicle, and $\ distance_{RSS}$ is the minimum safe distance measured by RSS. We measure both values at each timestep of an episode. If  $\ distance_{Euclidean}$ < $\ distance_{RSS}$, then it is marked as a failure timestep using the reward function of the RL agent. We have awarded +2 to bias the RL agent towards creating more failure timesteps. Similarly, if  $\ distance_{Euclidean}$ >=  $\ distance_{RSS}$, then it is considered as a success timestep. A success timestep represented as $\ success_{ts}$. We consider the overall scenario as a success scenario if more than 75$\%$ of the timesteps are a success within an episode. Otherwise, it will be designated as a failure scenario. We arbitrarily chose 75$\%$ as a threshold limit to demonstrate the proposed method, further study into the appropriate values for this number are part of future work. The metric for evaluating the specification does not affect the training of the agent, though the number of failure timesteps is used to determine whether a particular scenario is a success or failure as per the specification below:
\begin{multline}
success_{scenario}=(\frac{1}{N}\sum_{i=1}^{N}success_{ts} )*100\ > 75\% 
\end{multline}

\textbf{Implementation}: 
The overall approach is to determine the statistical characteristics of the SUT's behavior in the experimental context using fewer scenarios compared to other approaches. As discussed, RL is employed to generate the scenarios for validation. To apply RL algorithms, we model the environment as a MDP. In this experiment, each state obeys the Markov property. In other words, the future state is independent of past states given the current state, and we can observe the states from the environment.

Firstly, this section explains how RL can learn to generate edge cases. The Carla simulator has a synchronous mode which enables the simulator to pause until the processing on the client-side finishes. This is essential as the RL algorithm needs to perform the action and observe the next state before inputting the next action. The pedestrian crossing intersection has been set up in Carla as an environment. In each episode, there will be N number of time steps.  In each time step$\ t$, the DQN agent performs the action, observes the new state and determines how good the action was using the reward function. In this experiment, at each time step the action is applied to the pedestrian in the form of modifying the speed, and the new state is observed. The state is represented using relative speed and euclidean distance of the pedestrians w.r.t ego vehicle. 

The number of timesteps in an episode varies based on how the episode was concluded.  Before starting each episode, the environment will be reset in order to position the pedestrian and ego vehicle in the initial state. At the first timestep, the current state is calculated using the relative position and relative speed. Next, the current state is passed to the DQN code to select the action, which determines the speed of the pedestrian. Initially, the action will be selected randomly until the $ \epsilon$ value reaches the $\ \epsilon_{min}$. Afterward, the prediction network is used to predict the action for a given state and the reward function calculates the RSS metric and the appropriate reward value. The replay memory queue stores the relevant values at each timestep. Once the minimum batch size is reached, one epoch training of prediction network is complete. The whole process repeats at each time step until one or more of the end conditions is satisfied, meaning the episode is completed. Before starting the next episode, the environment will be reset with a new initial state of ego vehicle and pedestrian. This whole process repeats until the maximum number of episodes has been reached. 

At the end of each episode, we count the number of timesteps marked as success. If the total count is greater than the threshold limit of 75$\%$, we consider this as a success scenario. Otherwise, it will be marked as a failure scenario. While the RL system is in exploration mode, there will be significant variance in the scenarios, including some edge cases, which can be used to validate the statistical characteristics of the SUT's behavior. By biasing the learning towards the edge case, this method enables statistical validation with a fewer number of scenarios compared to validating against a large number of randomly generated scenarios.

\section{Results}
\begin{figure}[h]
\includegraphics[width=0.95\columnwidth, trim={0.5cm 0 2.1cm 0},clip ]{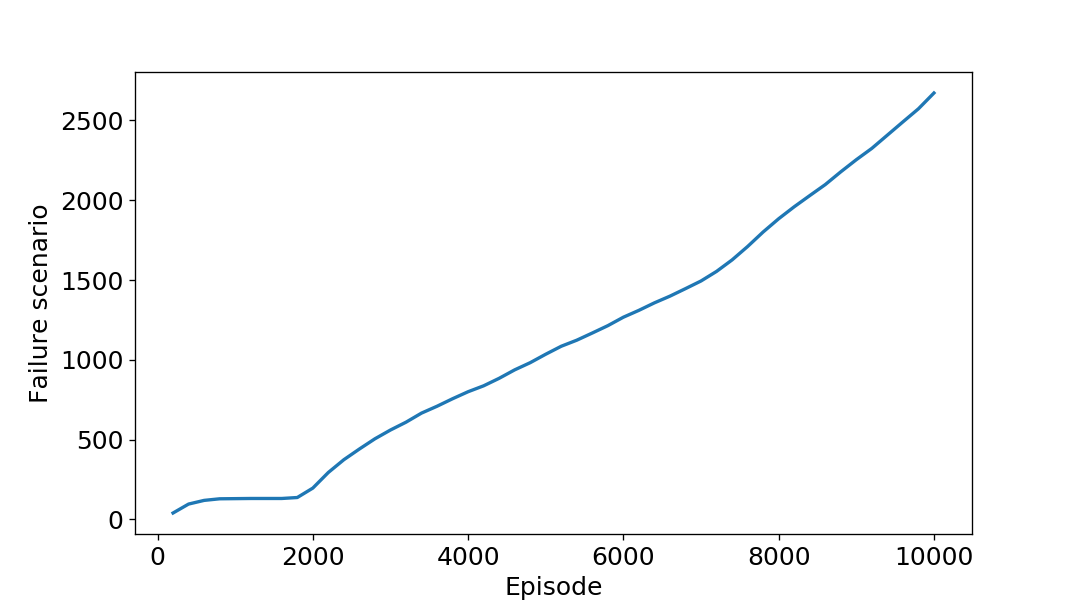}
\caption{ Total number of failure scenarios over 10,000 scenarios/episodes.}
\label{fig:failure_scenarios}
\end{figure}

The results of our approach are demonstrated using a number of metrics that show how the potentially unsafe edge cases can be learnt using RL.
Figure~\ref{fig:failure_scenarios} shows the accumulated number of failure scenarios for the system as it is trained over 10000 episodes. This figure shows that there are 2723 failure scenarios and 7277 successful scenarios over 10000 scenarios/episodes.

\begin{figure}[h]
\includegraphics[width=0.9\columnwidth, trim={0.5cm 0 2.1cm 0},clip]{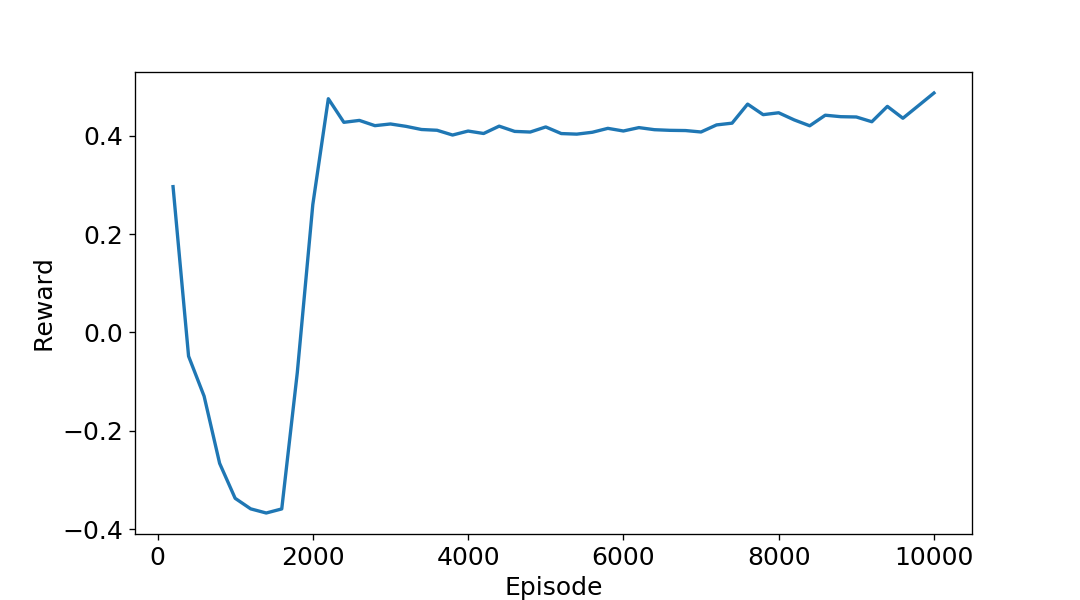}
\caption{It is a metric used to measure the performance of RL. The plot shows that after 2000 episodes, RL reward is almost constant. It indicates that RL has started to predict the right action for a given state that maximises the reward.}
\label{fig:reward_per_episode}
\end{figure}

The probability of SUT meeting the specification as defined in this approach is calculated by dividing the total number of success scenarios by the total count of scenarios.
\begin{equation}
P_r = \frac{Success_{scenario}}{Total_{scenario}} = \frac{7277}{10000} = 72\%
\end{equation}
In this experiment, the SUT has met the system specification requirements 72$\%$ of the time. 
The proposed method enabled us test the system against challenging cases using far fewer scenarios compared to randomised inputs. 
As shown Figure~\ref{fig:reward_per_episode}, the RL agent started to converge after just over 2000 episodes.  The average reward remains almost constant after this point. It means that the agent learned to predict the correct action that maximizes the failure cases, or in other words, the agent learned to create a high proportion of edge cases. This reduces the need for billions of miles of data or a large number of randomly generated scenarios to find a significant number of challenging scenarios for the SUT.

During the initial stages, random actions can create failure scenarios as illustrated in Figure~\ref{fig:failure_scenarios}. This can be interpreted as stress testing using randomly generated test cases.

\begin{figure}[ht]
\begin{subfigure}{\columnwidth}
  \centering
  \includegraphics[width=0.95\columnwidth, trim={2.9cm 0 1.7cm 0},clip]{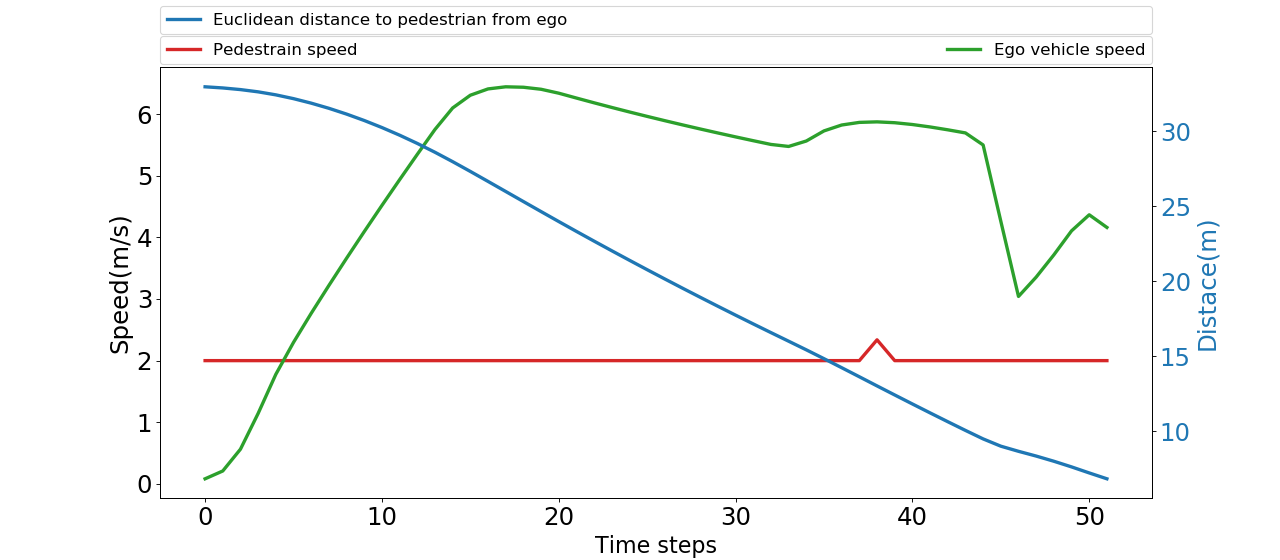}  
  \caption{An edge case with less than 75\% success timesteps. It means that there are more than 25\% failure timesteps. The number of failure timesteps of this scenario can be identified from Figure~\ref{fig:challenging_scenario_reward}, where failure timesteps start to occur from timestep 38.}
  \label{fig:challenging_scenario}
\end{subfigure}

\begin{subfigure}{\columnwidth}
  \centering
  \includegraphics[width=0.9\columnwidth, trim={0.5cm 0 2.1cm 0},clip]{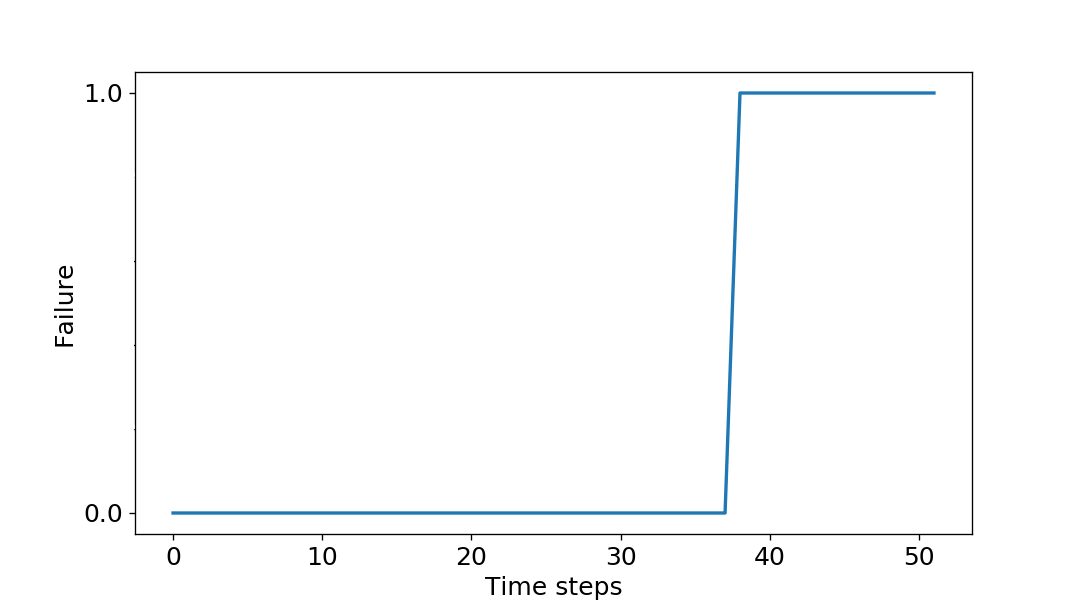}  
  \caption{Number of failure timesteps in the failure scenario as shown in the Figure~\ref{fig:challenging_scenario}. It indicates that failure timesteps start to occur from timestep 38 onwards.}
  \label{fig:challenging_scenario_reward}
\end{subfigure}
\caption{This plot shows speed, euclidean distance, and the number of failure timesteps required to create the edge case.}
\label{fig:fig}
\end{figure}

After 2000 episodes, the RL agent converged to a solution containing a high proportion of edge cases. A typical unsafe scenario with high reward value is illustrated in Figure~\ref{fig:challenging_scenario}. The diagram depicts the euclidean distance from the vehicle to the pedestrian(blue line), pedestrian speed(red), and ego vehicle speed(green) at each time step of an episode that was awarded the maximum reward (high number of failure cases). This experiment is done without considering uncertainties in the environment or sensor measurement, which could have an effect on the result. Figure~\ref{fig:challenging_scenario} and Figure~\ref{fig:challenging_scenario_reward} show an example of a scenario where the metric has indicated a number of timesteps where the system has failed. The pedestrian speed of 2 m/s is a contributing factor resulting in the pedestrian being in front of the vehicle at a critical time when the ego vehicle is close.
Failure cases occur from timestep number 38 onwards, as shown in Figure~\ref{fig:challenging_scenario_reward}. 

\begin{figure}[ht]
\vspace{3mm}
\begin{subfigure}{\columnwidth}
  \centering
  \includegraphics[width=\columnwidth, trim={0.5cm 0 2.1cm 0},clip]{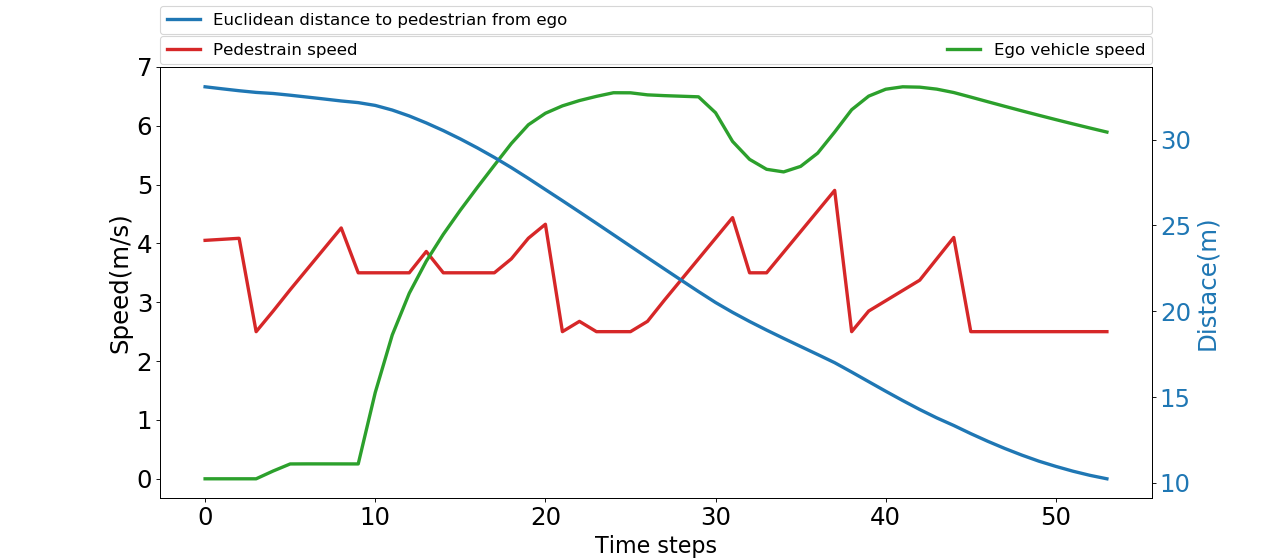}  
  \caption{A success scenario with more than 75\% of timesteps in the scenario marked as successful. As shown in Figure~\ref{fig:success_scenario_reward}, there is only one timestep in this scenario which is marked as a failure.}
  \label{fig:success_scenario}
\end{subfigure}
\begin{subfigure}{\columnwidth}
  \centering
  \includegraphics[width=0.9\columnwidth, trim={0.5cm 0 2.1cm 0},clip]{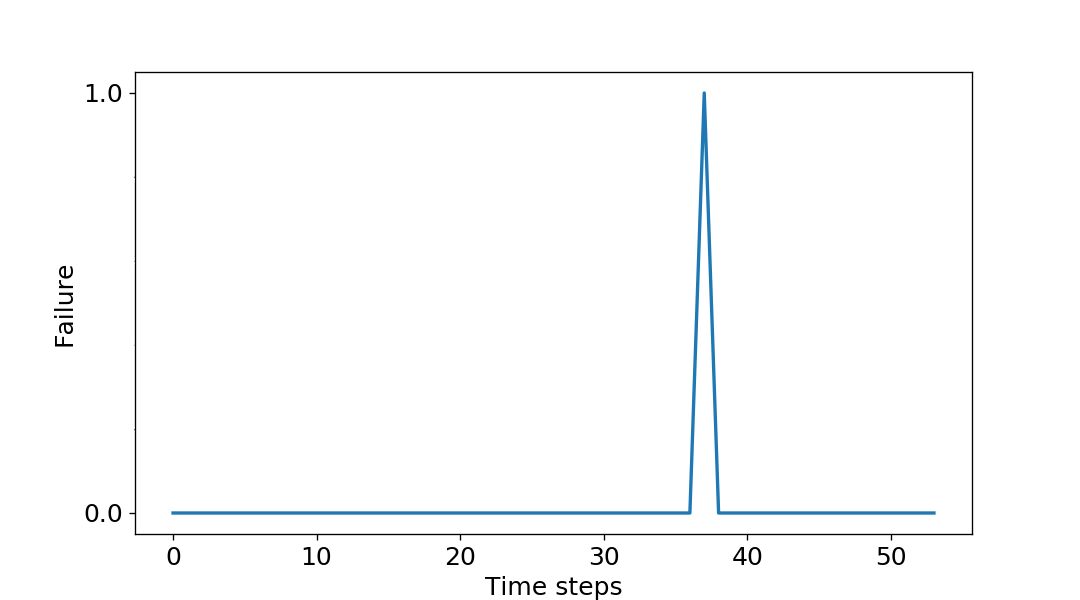}  
  \caption{Number of failure timesteps in the success scenario as shown in Figure~\ref{fig:success_scenario}.}
  \label{fig:success_scenario_reward}
\end{subfigure}
\caption{This plot shows speed, euclidean distance, and the number of failure timesteps of a typical success scenario}
\label{fig:fig}
\end{figure}

Figure~\ref{fig:success_scenario} and Figure~\ref{fig:success_scenario_reward} show an example of a scenario where the safety constraints are violated only for one time step.
The number of failure timesteps are considerably fewer resulting in a  success scenario. 
This is due to the pedestrian speed being much higher (around 4 m/s) meaning that the pedestrian has already crossed the road before the vehicle is in the critical area.
For this simplified experiment, this result aligns will human intuition. The failure cases occur when the pedestrian is walking at a speed that will cause them to be on the road when the vehicle is at high speed and in a critical location. 
The cases that were successful included scenarios where the pedestrian walked significantly faster, meaning that they had already crossed the road when the vehicle reached the critical location.

\section{Discussion and Conclusion}

This paper presented a new approach to compute the statistical characteristics of a system's behavior by biasing automatically generated test cases towards unsafe scenarios based on a safety metric. It enabled us to validate the system under test (SUT) and satisfy the specifications with a fewer number of scenarios.

In this work, we deal with the problem of non-determinism in testing by enabling statistical validation against thousands of generated scenarios with a focus on finding edge cases. For the experiment, we consider a pedestrian crossing and a collision avoidance system (CAS) as the SUT to be validated. Our result indicates that this method is an effective approach to do statistical validation since it can learn significant safety edge cases without relying on extensive on-road testing, which is expensive and time-consuming. Furthermore, biasing towards edge cases reduces the number of scenarios required to enable the statistical validation in a simulation environment.

The experimental results have shown that the RL agent can generate statistically meaningful scenarios for a SUT in order to find safety edge cases. This is done with fewer scenarios compared to other approaches. In this simplified experiment, we can also compare the result with human intuition.  In this case, most failure cases are created when the pedestrians are at the center of the road when the ego-vehicle is close and driving at maximum speed. A more complex scenario should present edge cases that may not be immediately intuitive, but will be automatically generated by the approach presented.

In future work, we will be increasing the complexity of the experiments by including more parameters for the RL agent to learn, and compare against other variants of CAS. Our automated vehicle platforms at USYD have implemented a sophisticated CAS that can be simulated and connected to the ego vehicle in the CARLA simulator. This ROS based system requires LIDAR point cloud and odometry. Once this system is connected to the simulator, it will receive this information from the simulator to observe the obstacles and determine the required speed at each time step. The methods presented will then be used to adjust the parameters of the USYD CAS to satisfy the RSS constraints in a variety of high complexity scenarios. 
Furthermore, we are extending our research to accommodate uncertainties in the learning process, such as adding sensor noise and pedestrian intentions.

\section*{ACKNOWLEDGMENT}

This work has been funded by the Australian Centre for Field Robotics (ACFR), University of Sydney and Insurance Australia Group (IAG) and iMOVE CRC and supported by the Cooperative Research Centres program, an Australian Government initiative
.

\bibliographystyle{IEEEtran}
\bibliography{references.bib}

\end{document}